\documentclass[sigconf]{acmart}
\usepackage{algorithm,algorithmic}
\usepackage{enumerate}
\usepackage{array}
\usepackage{subcaption}
\usepackage[skip=4pt]{caption}

\usepackage{scalerel}
\AtBeginDocument{%
  \providecommand\BibTeX{{%
    \normalfont B\kern-0.5em{\scshape i\kern-0.25em b}\kern-0.8em\TeX}}}

\setcopyright{rightsretained}

\acmSubmissionID{1579}

\begin{document}

\title{Adversarial Knowledge Transfer from Unlabeled Data}

\author{Akash Gupta}
\authornote{Joint first authors \\ ${}^{1}$Images Source: \url{www.pixabay.com}}
%
\affiliation{%
  \institution{University of California, Riverside}
}
\email{agupt013@ucr.edu}

\author{Rameswar Panda}
\authornotemark[1]
\affiliation{%
  \institution{MIT-IBM Watson AI Lab}
}
\email{rpand002@ucr.edu}

\author{Sujoy Paul}
\affiliation{%
  \institution{University of California, Riverside}
}
\email{spaul003@ucr.edu}

\author{Jianming Zhang}
\affiliation{%
  \institution{Adobe Research}
}
\email{jianmzha@adobe.com}

\author{Amit K. Roy-Chowdhury}
\affiliation{%
  \institution{University of California, Riverside}
  }
\email{amitrc@ece.ucr.edu}

\begin{abstract}
  While machine learning approaches to visual recognition offer great promise, most of the existing methods rely heavily on the availability of large quantities of labeled training data. However, in the vast majority of real-world settings, manually collecting such large labeled datasets is infeasible due to the cost of labeling data or the paucity of data in a given domain. In this paper, we present a novel \textbf{A}dversarial \textbf{K}nowledge \textbf{T}ransfer (\textbf{AKT}) framework for transferring knowledge from internet-scale unlabeled data to improve the performance of a classifier on a given visual recognition task.  
  The proposed adversarial learning framework aligns the feature space of the unlabeled source data with the labeled target data such that the target classifier can be used to predict pseudo labels on the source data. 
  An important novel aspect of our method is that the unlabeled source data can be of different classes from those of the labeled target data, and there is no need to define a separate pretext task, unlike some existing approaches. Extensive experiments well demonstrate that models learned using our approach hold a lot of promise across a variety of visual recognition tasks on multiple standard datasets. Project page is at \href{https://agupt013.github.io/akt.html}{\texttt{https://agupt013.github.io/akt.html}}. 
\end{abstract}

\ccsdesc[500]{Computing methodologies~Transfer learning}
\ccsdesc[500]{Computing methodologies~Image representations}

\keywords{Adversarial Learning, Knowledge Transfer, Feature Alignment.}

\copyrightyear{2020}
\acmYear{2020}
\acmConference[MM '20]{Proceedings of the 28th ACM International
Conference on Multimedia}{October 12--16, 2020}{Seattle, WA, USA}
\acmBooktitle{Proceedings of the 28th ACM International Conference on
Multimedia (MM '20), October 12--16, 2020, Seattle, WA, USA}
\acmDOI{10.1145/3394171.3413688}
\acmISBN{978-1-4503-7988-5/20/10}

\maketitle

\section{Introduction}
Deep learning approaches have recently shown impressive performance on many visual tasks by leveraging large collections of labeled data. 
However, such strong performance is achieved at a cost of creating these large datasets, which typically requires a great deal of human effort to manually label samples. 

\begin{figure}
\vspace{0.5em}
\centering
\begin{subfigure}[b]{0.48\textwidth}
    \centering
    \includegraphics[width=0.95\linewidth]{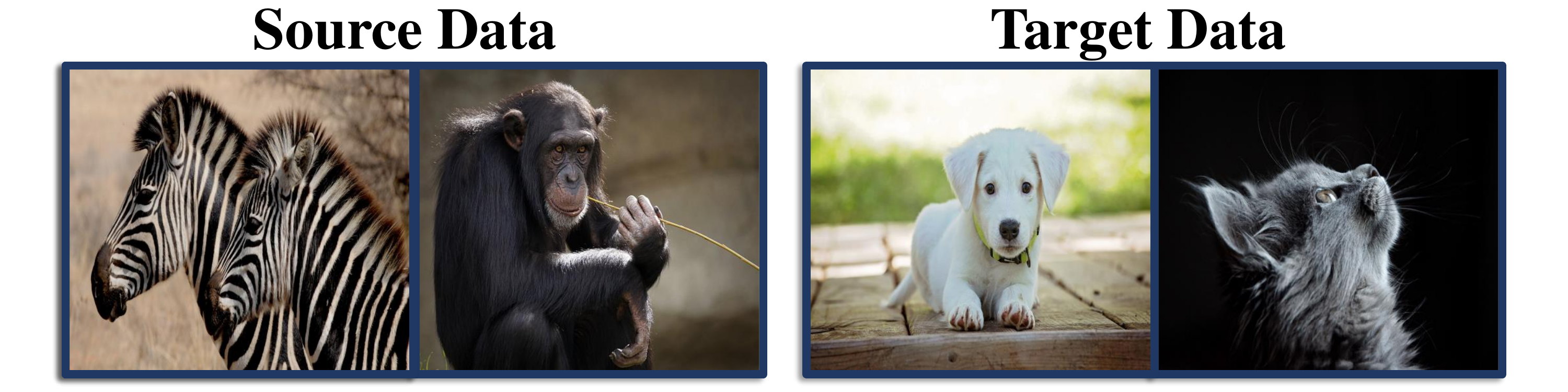}
    \vspace{-0.6em}
    \caption{Transfer Learning (Source: Labeled, Target: Labeled)}
    \label{fig:compare_a}
\end{subfigure}

\begin{subfigure}[b]{0.48\textwidth}
    \centering
    \includegraphics[width=0.95\linewidth]{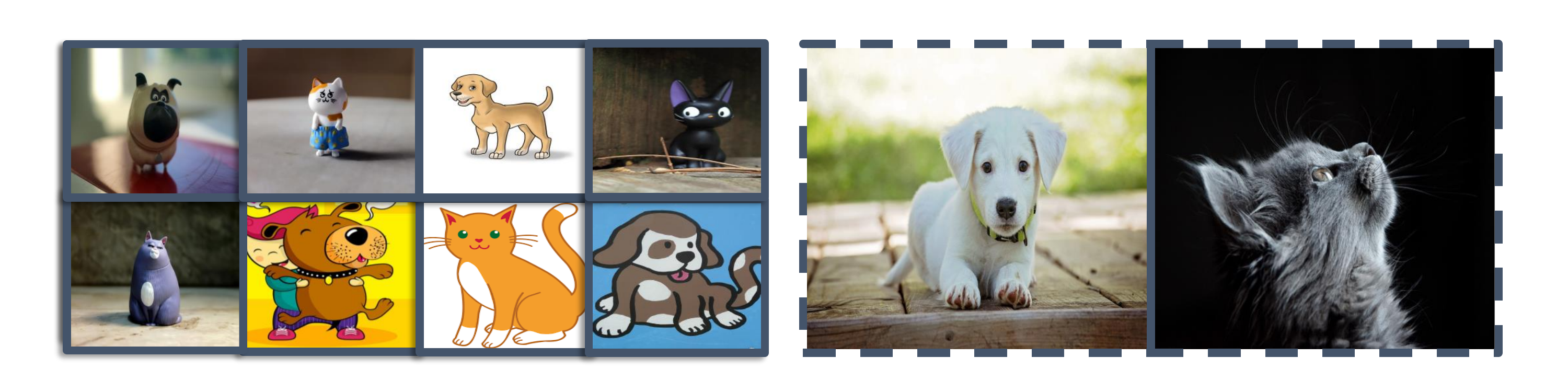}
    \vspace{-0.9em}
    \caption{Unsupervised Domain Adaptation (Source: Labeled, Target: Unlabeled)}
    \label{fig:compare_b}
\end{subfigure}

\begin{subfigure}[b]{0.48\textwidth}
    \centering
    \includegraphics[width=0.95\linewidth]{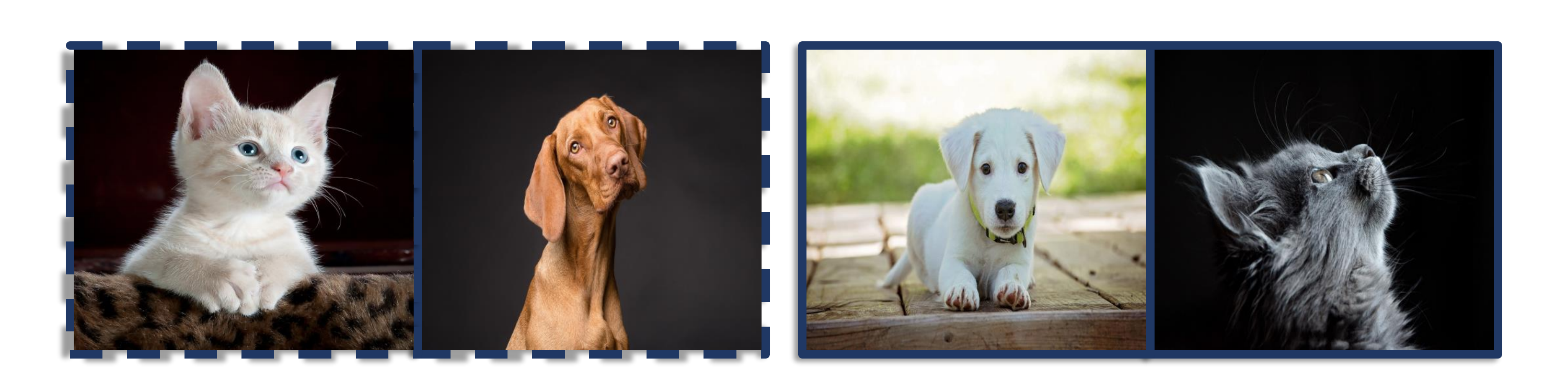}
    \vspace{-0.9em}
    \caption{Semi-Supervised Learning (Source: Unlabeled, Target: Labeled)}
    
    \label{fig:compare_c}
\end{subfigure}

\begin{subfigure}[b]{0.48\textwidth}
    \centering
    \includegraphics[width=0.95\linewidth]{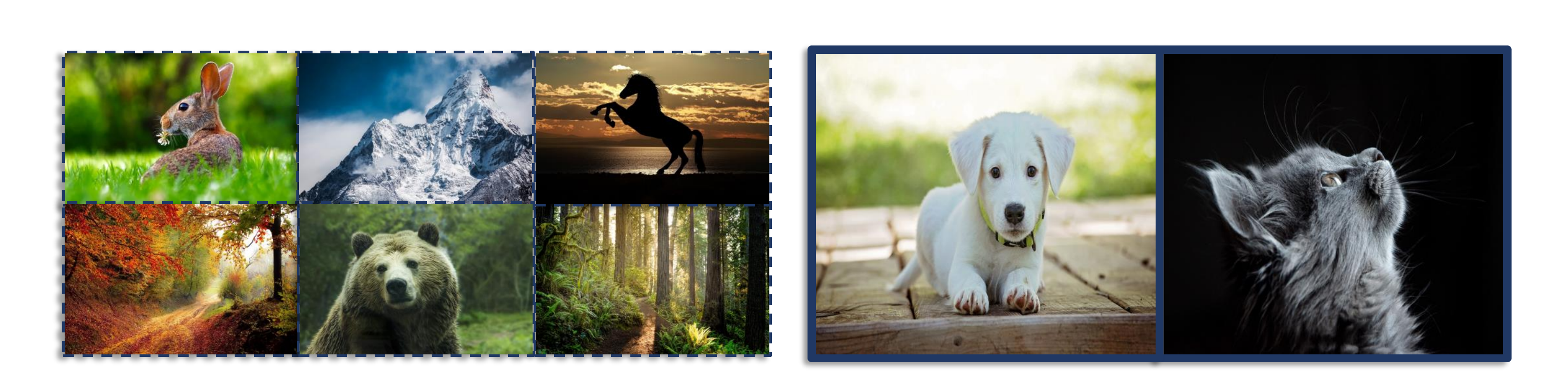}
    \vspace{-0.9em}
    \caption{AKT (Ours) (Source: Unlabeled, Target: Labeled)}
    \label{fig:compare_d}
\end{subfigure}
\caption{Comparison of different learning paradigms. Our proposed approach Adversarial Knowledge Transfer (AKT) transfers knowledge from unlabeled source data to the labeled target dataset without requiring them to have the same distribution or label space. Best viewed in color.$^1$
}
\label{fig:compare} 
\end{figure}

Recently, much progress has been made in developing a variety ways such as transfer learning~\cite{long2017deep,oquab2014learning,xu2017unified}, unsupervised domain adaptation~\cite{ganin2016domain, long2013transfer, long2016unsupervised} and semi-supervised learning ~\cite{grandvalet2005semi,laine2016temporal,sajjadi2016mutual,tarvainen2017mean,belkin2006manifold,ando2007learning,zhu2002learning,sajjadi2016regularization,miyato2018virtual,tarvainen2017mean} to overcome the lack of labeled data in target domain.
While these approaches have shown to be effective in many tasks, they often depend on existence of large scale annotated data to train a source model~\cite{yosinski2014transferable} (in case of transfer learning: Fig.~\ref{fig:compare_a}) or assume that unlabeled data comes from a similar distribution with some domain gap (e.g., across animated and real images in case of unsupervised domain adaptation: Fig.~\ref{fig:compare_b}) or have the same data and label distribution as of the labeled data (in case of semi-supervised learning: Fig.~\ref{fig:compare_c}). 
%
%
%
On the other hand, self-taught learning or self-supervised learning~\cite{raina2007self,lee2009exponential, dai2008self, wang2013robust, hou2014domain,noroozi2018boosting} methods can transfer knowledge to the target task using unlabeled source data, which may not have the same distribution and the same label space as that of the target task.
In particular, unlabeled data is first used to learn feature representation using a pretext task and then the learned feature is adapted to the target labeled dataset through finetuning. 
However, despite their reasonable performance, it is still unclear how to design efficient pretext tasks for specific downstream tasks. Defining a pretext task is a challenging problem on its own merit~\cite{kolesnikov2019revisiting}. 
%


In this paper, we propose a novel \textbf{A}dversarial \textbf{K}nowledge \textbf{T}ransfer (\textbf{AKT}) framework for transferring knowledge from large-scale unlabeled data \emph{without the need to define pretext tasks}. 
Our approach transfers knowledge from unlabeled source data to the labeled target data without requiring them to have the same distribution or label space (see Fig.~\ref{fig:compare_d}).
The proposed adversarial learning helps to align the feature space of unlabeled source data with labeled target data such that the target classifier can be used to predict pseudo-labels on source data. Using the pseudo labels of source data, we jointly optimize the classifier with the labeled target dataset. 
Unlike other self-supervised methods~\cite{zhang2016colorful,noroozi2018boosting},
we do not employ two stage training where a model is first trained on a pretext task and then finetuned on target task. 
Instead, our method operates just like a solver, which updates the model with labeled and unlabeled data simultaneously, making it highly efficient and convenient to use. 

Our approach is motivated by the observation that many randomly downloaded unlabeled images (e.g., web images from other object classes--which are much easier to obtain than images specifically of the target classes) contain basic visual patterns (such as edges, corners) that are similar to those in labeled images of the target dataset. Thus, we can transfer these visual patterns from the internet-scale unlabeled data for learning an efficient classifier on the target task. 
Note that the source and target domains in our approach may have some relationship, but we do not require them to have the same distribution or label space while transferring knowledge from the unlabeled data.

\begin{figure*}[ht] 
\vspace{1mm}
\centering
\includegraphics[scale=0.22]{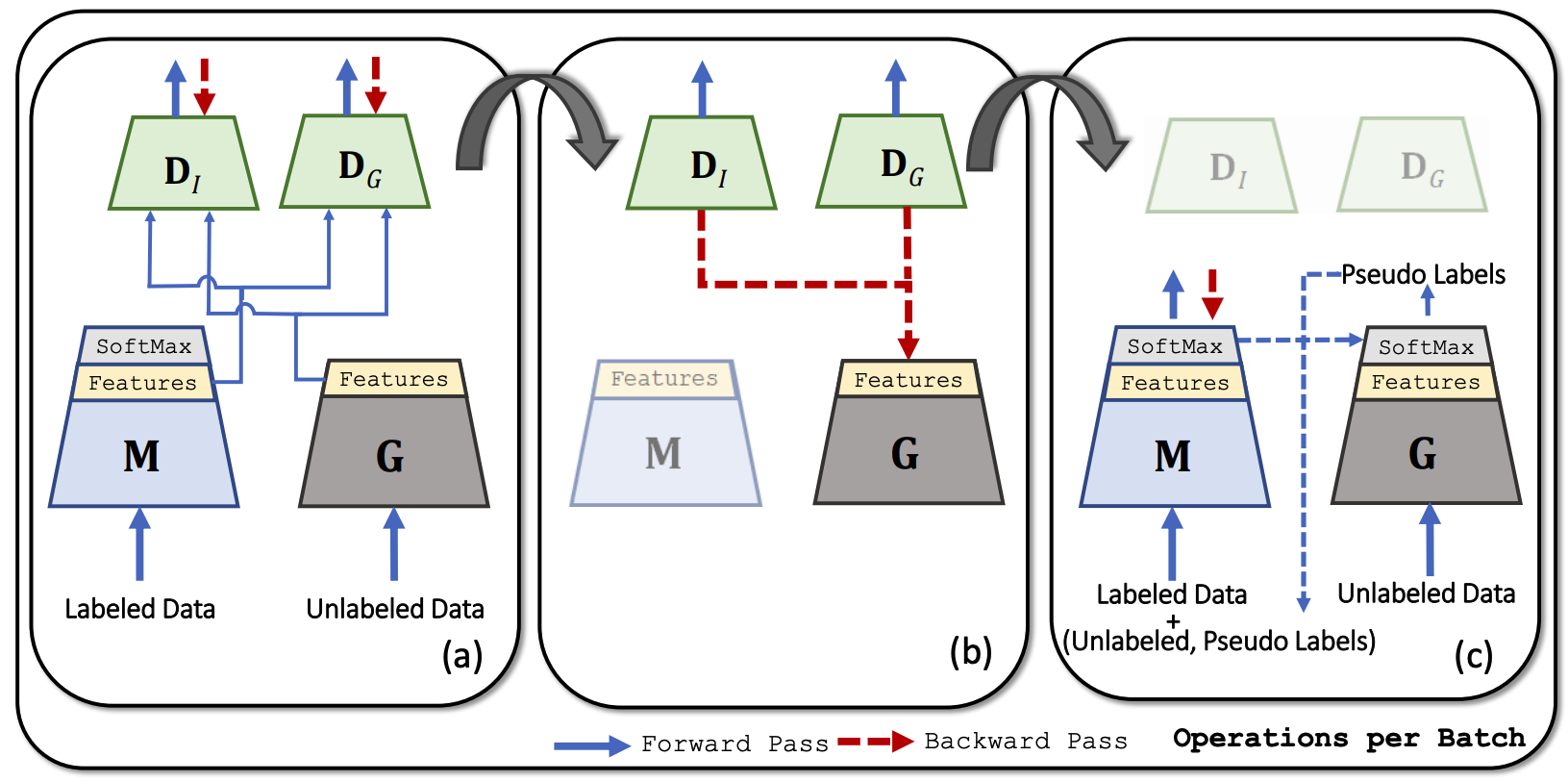}\vspace{1mm}
\caption{Overview of different operations occurring per batch in the proposed approach.
Given a small target dataset, with true labels, and a large amount of unlabeled source data, (a) we first forward pass both labeled and unlabeled samples through the classifier $\mathbf{M}$ and the pseudo-label generator $\mathbf{G}$, respectively, to extract their features. We then train a instance-level discriminator $\mathbf{D}_{I}$ to distinguish between target and source features for each instance and a group-level discriminator $\mathbf{D}_{G}$ to distinguish between mean of target and source features in a batch. (b) Next, we update $\mathbf{G}$ to fool both $\mathbf{D}_{I}$ and $\mathbf{D}_{G}$ using adversarial loss. 
(c) We copy weights of softmax layer from $\mathbf{M}$ to $\mathbf{G}$ and generate pseudo-labels for unlabeled samples. We then update the classifier $\mathbf{M}$, which is the main output of our approach, using both labeled target and pseudo-labeled source data. Best viewed in color.
}
\label{fig:method} 
\end{figure*}

\subsection{Approach Overview}
An overview of our approach is illustrated in Figure~\ref{fig:method}. Given a small labeled target data and large-scale unlabeled source data, our objective is to boost the performance of the target classifier by exploiting relevant information from unlabeled samples as compared to using only the labeled target data.
To this end,
we adopt an adversarial approach to train the target classifier that consists of three modules: the classifier network $\mathbf{M}$, the pseudo label generator $\mathbf{G}$ and the discriminators, $\mathbf{D}_I$ for instance-level feature alignment and $\mathbf{D}_G$ for group-level feature alignment. Note that learning the classifier $\mathbf{M}$ is our main goal. 

In our approach, to train the target classifier, (1) we first input the target samples to the classifier network $\mathbf{M}$ to extract the features.
(2) We then input the source samples to the pseudo label generator $\mathbf{G}$ to extract their features and pseudo-labels. (3) We use features of the target and source samples as input to both the discriminators $\mathbf{D}_I$ and $\mathbf{D}_G$ to distinguish whether the instance-level and group-level input features are from the source or target domain respectively. While instance-level discriminator tries to align individual samples from source domain to the target domain, the group level discriminator aligns one batch of source samples with one batch of target samples by considering feature means.
With these two adversarial losses, the network propagates gradients from $\mathbf{D}_I$ and $\mathbf{D}_G$ to $\mathbf{G}$, which encourages generation of similar feature distributions of source samples in the target domain.
(4) Finally, we pass target samples and source samples to the classifier $\mathbf{M}$ to compute the classification loss on target samples with their labels and source samples with the pseudo labels generated from the pseudo-label generator $\mathbf{G}$.

\subsection{Contributions}
To summarize, we address a novel and practical problem in this paper - how to leverage information contained in the unlabeled data that does not follow the same class labels or generative distribution as the labeled data. Towards solving this problem, we make the following contributions.\medskip

\noindent(1) We propose a novel adversarial framework for transferring knowledge from unlabeled data to the labeled data without any explicit pretext task, making it highly efficient.\medskip

\noindent (2) We perform extensive experiments on multiple datasets and show that our method achieves promising results, while comparing with state-of-the-art methods, without requiring any labeled data from the source domain.

\section{Related Work}
Our work relates to four major research directions: semi-supervised learning, domain adaptation, self-training and  self-taught learning.\medskip

\textbf{Semi-Supervised Learning} has been studied from multiple perspectives (see review~\cite{chapelle2009semi}). A number of prior works focus on adding regularizers to the model training which prevent overfitting to the labeled examples~\cite{grandvalet2005semi,laine2016temporal,sajjadi2016mutual,tarvainen2017mean}. Various strategies have been studied, including manifold regularization~\cite{belkin2006manifold}, regularization with graphical constraints~\cite{ando2007learning}, and label propagation \cite{zhu2002learning}. In the context of deep neural networks, semi-supervised learning has also been extensively studied using ladder networks~\cite{belkin2006manifold}, temporal ensembling~\cite{laine2016temporal}, stochastic transformations~\cite{sajjadi2016regularization}, virtual adversarial training~\cite{miyato2018virtual}, and mean teacher~\cite{tarvainen2017mean}. While most of the existing semi-supervised methods consider the unlabeled set to have the same distribution as the labeled set, we do not assume any correlation between unlabeled data and classification tasks of interest.\medskip

\textbf{Domain Adaptation} aims to transfer knowledge from the labeled source data to the unlabeled target data assuming that both contain exactly the same number of classes ~\cite{ganin2016domain, long2013transfer, long2016unsupervised}. In contrast, our work focuses on the opposite case of knowledge transfer without any assumption on the label space. Though open-set domain adaptation~\cite{panareda2017open, saito2018open, liu2019separate} does not consider exactly the same classes, but they still have a few classes of interest that are shared between source and target data. In contrast, we don't require them to have the same distribution or shared label space while transferring knowledge from unlabeled data.\medskip

\textbf{Self-Training} is a learning paradigm where information from a small set of labeled data is exploited to estimate the pseudo labels of unlabeled data. 
Pseudo labeling~\cite{lee2013pseudo, bachman2014learning} is a commonly used technique where a model trained on the label set is first used to predict the pseudo labels of unlabeled set and jointly trained with both labeled and unlabeled data. However, this may result in incorrect labeling if the initial model learned from labeled instances is overfitted. Moreover, it makes an implicit assumption that the source sample features are well aligned with the target data which may not be true. In contrast, our approach makes no such assumption and trains a generator to produce more reliable pseudo-labels by aligning source sample features with the target data distribution.\bigskip

\textbf{Self-Taught Learning or Self-Supervised Learning} mostly defines a pretext task to learn the feature representation from a large unlabeled set and then finetunes the learned model on the target task involving a much smaller labeled set. 
Some methods define the pretext task as denoising auto-encoder where the task is data reconstruction at the pixel level from partial observations~\cite{vincent2008extracting}. 
The colorization problem~\cite{zhang2016colorful,larsson2016learning} is also a notable example, where the task is to reconstruct a color image given its gray scale version. Image inpainting~\cite{pathak2016context} is also used as a pretext task, where the goal is to predict a region of the image given the surrounding. Solving Jigsaw puzzles~\cite{noroozi2016unsupervised} or its variant~\cite{noroozi2018boosting} has been used as pretext task for learning visual features from unlabeled data. However, defining pretext tasks for certain applications is a challenging problem on its own merit~\cite{kolesnikov2019revisiting}.
%
While our approach is related to self-taught learning, unlike existing works that often employ two stage training, we exploit unlabeled data along with labeled data using an end-to-end framework which updates the classifier with labeled and unlabeled data simultaneously, making it highly efficient.

\section{Methodology}\label{sec:method}
We propose an \textbf{A}dversarial \textbf{K}nowledge \textbf{T}ransfer (\textbf{AKT}) framework for transferring knowledge from unlabeled to labeled data without requiring any correspondence across the label spaces. Our goal is to leverage unlabeled images from other object classes which are much easier to obtain than images of the target classes to improve the test accuracy on the target task. We first precisely define the problem that we aim to solve in this work and then present our knowledge transfer approach followed by the optimization details.

\subsection{Problem Statement} 
Consider a labeled dataset $\mathcal{X}_t=\{(x_t^1, y_t^1 ), (x_t^2,y_t^2), \cdots ,(x_t^n,y_t^n)\}$ and an unlabeled source dataset $\mathcal{X}_s=\{x_s^1,x_s^2, \dots, x_s^m\}$, where $y_t^i \in {\rm I\!R}^c$ is the class label of target sample $\mathbf{x}_t^i$, $c$ is the total number of target classes and $m >> n$. Our objective is to generate pseudo-labels $\mathbf{\bar{y}}_s=\{\bar{y}_s^1, \bar{y}_s^2, \dots,\bar{y}_s^m\}$ for unlabeled samples in the label space of the target domain such that it improves the performance of the classifier  when trained jointly. More specifically, we aim to use the unlabeled set $\mathcal{X}_s$ along with $\mathcal{X}_t$ to boost the performance of the classifier compared to using only $\mathcal{X}_t$.

\subsection{Adversarial Knowledge Transfer}
Let us define the classifier $\mathbf{M}(x)$, $x \in \mathcal{X}_t \cup \mathcal{X}_s$ using a deep convolutional neural network and the pseudo-label generator $\mathbf{G}(x)$, $x \in \mathcal{X}_s$ with the same architecture as the classifier. The goal of pseudo-label generator is to predict the labels of the unlabeled source samples, $\mathbf{\bar{y}}_s$ in such a way so that the source samples seem to have been drawn from the labeled data distribution. However, since the unlabeled source samples do not follow the same label space as the target data, we need feature alignment across unlabeled and labeled samples such that more reliable pseudo-labels 
can be generated for the unlabeled source samples. 
To achieve this, we introduce two discriminators, namely instance-level and group-level discriminator which act as feature aligners between the classifier,$\mathbf{M}$, and the pseudo-label generator, $\mathbf{G}$. The instance-level discriminator, $\mathbf{D}_{I}$, learns to detect whether the feature is from the classifier or pseudo-label generator using an adversarial training. On the other hand, the group level-discriminator, $\mathbf{D}_{G}$, aims to learn holistic statistical information by distinguishing if the mean feature is from the classifier or the pseudo-label generator.
Specifically, through these two discriminators, we try to discover both localized knowledge at the instance level and the global feature distribution knowledge while aligning features across both source and target domains.

Let the feature representation at $k^{th}$ layer be defined as $\mathbf{M}^k(x)$ and $\mathbf{G}^k(x)$ for the classifier and the pseudo-label generator respectively. In our experiments, we choose the features from the last fully connected layer (denoted by $L$) as it has shown to be effective in many transfer learning tasks. 
We learn the parameters of $\mathbf{G}$, $\mathbf{D}_I$, $\mathbf{D}_G$ and $\mathbf{M}$ using adversarial training, as described in Algorithm ~\ref{alg:algo}. 
The learned classifier is finally used in evaluation on the target task.   
Moreover, we adopt the same network architecture for both, the classifier and the pseudo-label generator, in order to ensure that the feature representations of both labeled and unlabeled samples are in the same space for feature alignment. Notice that our pseudo-label generator mimics the generator in Generative Adversarial Networks (GANs)~\cite{goodfellow2014generative} which generates an image from a random vector. However, unlike generators in GANs, the input to our generator is an image rather than a latent noise vector and it produces pseudo-labels for the given inputs.

\subsection{Optimization}
For transferring knowledge from unlabeled data, we propose to adapt the two-player min-max game based on the discriminators and the pseudo-label generator. Towards this, we need to learn the parameters of generator $\mathbf{G}$, and the discriminators in a way that $\mathbf{G}$ is able to generate features such that both $\mathbf{D}_{I}$ and $\mathbf{D}_{G}$ are not able to discriminate between instance-level and group-level features from target or source distribution respectively. We consider classifier features as positive and pseudo-label generator features as negative and then train the discriminators using binary cross-entropy loss. On the other hand, the generator is trained to produce features such that the discriminators are fooled.
Thus, similar to the GANs, the pseudo-label generator $\mathbf{G}$ and the discriminators, $\mathbf{D}_I$ and $\mathbf{D}_G$, play the following two-player min-max game:
\begin{alignat}{2}\label{eqn:gan}
        \min_{\mathbf{G}}\max_{\mathbf{D}_I, \mathbf{D}_G} & 
         \mathbb{E}_{\scaleto{p_{M(x_t)}}{5pt}}
         \Bigg[
         \log  \mathbf{D}_I \Big(
                                \mathbf{M}^L(x_t) 
                            \Big)
                            +
         \log \mathbf{D}_G \Big(
                                \mathbf{M}^L(x_t)
                            \Big)
         \Bigg] ~+
         \\
        &\mathbb{E}_{
       \scaleto{p_{G(x_s)}}{5pt}}
        \Bigg[
        \log \bigg(
                    1-\mathbf{D}_I \Big(
                                        \mathbf{G}^L(x_s)
                                    \Big)
            \bigg)
            +
        \log \bigg(
                    1-\mathbf{D}_G \Big(
                                        \mathbf{G}^L(x_s)
                                    \Big)
              \bigg)
        \Bigg] \nonumber
    \end{alignat}
where, $p_{M(x_t)}$ and $p_{G(x_s)}$ correspond to the feature distribution of $\mathbf{M}$ in target domain and $\mathbf{G}$ in source domain. We solve the optimization problem in eqn. \ref{eqn:gan} alternately using gradient descent where we once fix the parameters of the generator $\mathbf{G}$ and train the discriminators $\mathbf{D}_I$ and $\mathbf{D}_G$ and vice versa as described below.\medskip

\noindent \textbf{Discriminator Training.} Given two discriminators for instance-level and group-level feature alignment across the classifier $\mathbf{M}$ and the pseudo-label generator $\mathbf{G}$, during training, we first update the weights of the discriminators $\mathbf{D}_{I}$ and $\mathbf{D}_{G}$ with features coming from $\mathbf{M}$ as positive and that coming from $\mathbf{G}$ as negative. For a batch of size $b$, instance-level discriminator loss ($\mathcal{L}_{D_{I}}$) and group-level discriminator loss ($\mathcal{L}_{D_{G}}$) on mean features are defined as follows. 

    \noindent
    \begin{equation}
        \mathcal{L}_{D_{I}} = \frac{1}{b}\sum_{i=1}^{b} \log \mathbf{D}_I \big(\mathbf{M}^L(x_t^i) \big)
         + ~\frac{1}{b}\sum_{i=1}^{b} \log \Big( 1-\mathbf{D}_I \big(\mathbf{G}^L(x_s^i) \big) \Big) 
          \label{eqn:D_inst_eqn}
    \end{equation} 
    \noindent
    \begin{equation}
        \mathcal{L}_{D_{G}} = 
         \log \mathbf{D}_G\Big( 
         \frac{1}{b}\sum_{i=1}^{b}\mathbf{M}^L(x_t^i) 
         \Big)
         + ~ \log \bigg(1-\mathbf{D}_G \Big( \frac{1}{b}\sum_{i=1}^{b}\mathbf{G}^L(x_s^i) 
         \Big) \bigg)
         \label{eqn:D_grp_eqn}
    \end{equation}

\noindent Finally, the total loss for discriminators $\mathcal{L}_D$ is given as
\begin{equation}
         \mathcal{L}_D = \lambda_{D_I}\mathcal{L}_{D_{I}} + \ \lambda_{D_G} \mathcal{L}_{D_{G}}
         \label{eqn:D_eqn}
\end{equation} 
where, $\lambda_{D_I}$ and $\lambda_{D_G}$ are the weights for the instance and group discriminator losses respectively.\medskip

\begin{algorithm} [t]
   \caption{Adversarial Knowledge Transfer Training}
   \label{alg:algo}
\begin{algorithmic}
   \STATE {\bfseries Input:} $\mathcal{X}_t = \{(\boldsymbol{x_t}^i, y_t^i)\}_{i=1}^n$, $\mathcal{X}_s= \{\boldsymbol{x_s}^j\}_{j=1}^m$
   \STATE {\bfseries Output:} Classifier Model $\mathbf{M}$
   
   \FOR{number of training iterations}
   \STATE Sample $b$ tuples of $(\boldsymbol{x_t}, y)$ from $\mathcal{X}_t$ 
   \STATE Sample $b$ samples of $\boldsymbol{x_s}$ from $\mathcal{X}_s$
   \begin{enumerate}[\quad \bf Step a.\hspace{-4mm}]
       \item ~Minimize $\mathcal{L}_D$ using Eqn.~\ref{eqn:D_eqn} and update $\mathbf{D}_I$, $\mathbf{D}_G$
       \item ~Maximize $\mathcal{L}_G$ using Eqn.~\ref{eqn:G_eqn} and update $\mathbf{G}$
       \item ~Minimize $\mathcal{L}_M$ using Eqn.~\ref{eqn:M_eqn} and update $\mathbf{M}$
   \end{enumerate}
   \ENDFOR
\end{algorithmic}
\end{algorithm}

\noindent \textbf{Pseudo-Label Generator Training.} The objective of training $\mathbf{G}$ is to fool the discriminators such that the discriminator fails to distinguish whether the feature is coming from the classifier or pseudo-label generator. The loss function to train the $\mathbf{G}$ can be written as follows.

\fontsize{8.9}{8.9} \selectfont{
\begin{equation}
    \mathcal{L}_G = \frac{1}{b}\sum_{i=1}^{b}  \log \bigg( 1-\mathbf{D}_I \Big(\mathbf{G}^L(x_s^i)\Big)\bigg) 
    +  \log \bigg( 1-\mathbf{D}_G \Big(\frac{1}{b}\sum_{i=1}^{b} \mathbf{G}^L(x_s^i)\Big)\bigg)
     \label{eqn:G_eqn}
\end{equation}}
\normalsize

\noindent \textbf{Classifier Training.} The classifier is updated at the end of each iteration using both labeled and unlabeled data. We can update $\mathbf{M}$ easily for the labeled data but since there are no labels for the unlabeled source data, we use prediction from the generator as true labels to compute the loss. The loss function to update the classifier for an iteration with batch size $b$ can be represented as follows:
\begin{align}
   \mathcal{L}_M = \frac{1}{b} \sum_{i=0}^b \mathcal{L}\Big(\mathbf{M}(x_t^i), y_t^i\Big) + \lambda_s~ \mathcal{L}\Big(\mathbf{M}(x_s^i), \mathbf{G}(x_s^i)\Big)
   \label{eqn:M_eqn}
\end{align}
where $\mathcal{L}(p,q)$ is the categorical cross-entropy loss multi-class and binary cross-entropy loss for multi-label classification, $p$ as the true label and $q$ as the predicted label. The loss originating from the source samples is used as a regularizing term with parameter $\lambda_s$ so that it assists the learning of target task and not suppressing it.

\section{Experiments}
We perform rigorous experiments on different visual recognition tasks, such as Object Recognition (single-label and multi-label), Character Recognition (Font and Handwritten), and Sentiment Recognition to verify the effectiveness of our approach. Our primary objective is to transfer knowledge from unlabeled data to the classifier network such that the test accuracy on the target data significantly improves over the training from scratch (i.e., training with only target data without any knowledge transfer) and approaches as close to that of supervised knowledge transfer methods that uses the labeled source data. 
Note that we do not require the source and target domains to have the same distribution or label space across all our experiments.

\subsection{Implementation Details} All our implementations are based on PyTorch~\cite{paszke2017automatic}. We choose VGG-16~\cite{simonyan2014very} as the network architecture for both classifier and generator in all our experiments except CIFAR-10 and Pasca-VOC experiments. For CIFAR-10 experiment, to maintain the same network architecture as with other methods, we modify the last three fully connected layers to (512-512-10) from (4096-4096-10). In Pascal-VOC experiment we use AlexNet to compare with other state-of-the-art methods~\cite{krahenbuhl2015data,pathak2017learning,pathak2016context,krahenbuhl2015data,zhang2016colorful,zhang2017split,noroozi2016unsupervised}. We choose the same network architecture for both classifier and pseudo-label generator as the features generated by these networks should lie in the same space and thus it helps in the feature alignment. We use three fully-connected (FC) layers (512-256-128) as the discriminator network for CIFAR-10 experiment and (4096-1024 -512) for all other experiments. 
We use SGD with learning rate 0.01, 0.01 and 0.001 for the classifier, pseudo-label generator and discriminator, respectively. We use momentum of 0.9 and weight decay of 0.0005 while training the classifier. The learning rate of the classifier and generator is reduced by a factor of 0.1 after 75 epochs. All the loss weights are set to 1 and kept fixed for all experiments. We train our models with a batch size of 96 for all experiments except for PASCAL-VOC where we use batch of 20. In each iteration, the generator is updated once using unlabeled source dataset and the discriminator is updated twice using labeled target and unlabeled source dataset.

\subsection{Compared Methods}
We compare with several methods that fall into two main categories. (1) Supervised knowledge transfer methods that use labeled source data, such as Finetuning and Joint Training (i.e., multi-task learning). (2) Unsupervised knowledge transfer methods (a.k.a self-supervised methods) that leverage unlabeled source data, such as Random Network~\cite{zhang2016understanding}, Pseudo Labels~\cite{lee2013pseudo}, Jigsaw~\cite{noroozi2016unsupervised}, Colorization~\cite{zhang2016colorful} and Split-Brain Autoencoder~\cite{zhang2017split}. 
We additionally compare with the training from scratch baseline to show the effectiveness of our knowledge transfer approach for improving recognition accuracy on the target task. Below are the brief descriptions on the baselines.\medskip

\noindent\textbf{Finetuning and Joint Training.} In Finetuning, we first train a classifier using source data by assuming that the true labels are available during training and then perform an end-to-end finetuning on the target dataset. The Joint Learning baseline learns a shared representation using a single network on both source and target datasets. Since both of these methods use labeled source data, we call them as supervised knowledge transfer methods.\medskip  

\noindent\textbf{Random Network.} Following~\cite{zhang2016understanding,noroozi2018boosting}, we perform an experiment to obtain random pseudo labels for source data by clustering the features extracted from a randomly initialized network. We then train the classifier network using these pseudo-labels for source data and finetune it on the labeled target dataset.\medskip

\noindent\textbf{Pseudo Labels.} We first train a classifier on the labeled target data and obtain the pseudo labels for source data by making predictions using the classifier. We then update the network using both labeled target and pseudo-labeled source images to obtain an improved model for recognition on target task.\medskip

\noindent\textbf{Jigsaw~\cite{noroozi2016unsupervised}.} We use solving jigsaw puzzles as pretext task on unlabeled source dataset and then finetune on target task. We use same network (VGG-16) to make a fair comparison with our approach.\medskip

\noindent\textbf{Colorization~\cite{zhang2016colorful}.} Following~\cite{zhang2016colorful}, we use colorization (mapping from grayscale to color version of a photograph) as a form of self-supervised feature learning on unlabeled source data and then finetune on labeled target data. \medskip

\noindent \textbf{Split-Brain Autoencoder~\cite{zhang2017split}.} Split-Brain Autoencoder uses the cross-channel prediction to learn features on unlabeled data where one sub-network solves the problem of colorization (predicting a and b channels from the L channel in Lab colorspace), and the other perform the opposite (synthesizing L from a, b channels).

As character recognition task involves grayscale images, we perform colorization and split-brain autoencoder experiments on only object recognition and sentiment recognition tasks. We use publicly available code of both methods and set the parameters as recommended in the published works. 

\subsection{Object Recognition}
The goal of this experiment is to verify the effectiveness of our proposed adversarial approach in both single and multi-label object recognition while leveraging unlabeled data (from different classes as the target task) which are abundantly available from the web. 

\subsubsection{Single-label Object Recognition}
\quad We conduct this experiment using CIFAR-10 \cite{krizhevsky2009learning} as the labeled target dataset and CIFAR-100 \cite{krizhevsky2009learning} as the unlabeled source dataset. 
Both datasets contain 50,000 training images and 10,000 test images of size $32\times32$. We use the same train/test split from the original CIFAR-10 dataset for our experiments. Note that the classes in CIFAR-10 and CIFAR-100 are mutually exclusive. Table~\ref{table:task_1} shows results of different methods on CIFAR-10 dataset. From Table~\ref{table:task_1}, the following observations can be made: (1) The classifier trained using our approach outperforms all the unsupervised knowledge transfer (self-supervised) methods.
Among the alternatives, Split-Brain baseline is the most competitive. However, our approach still outperforms it by a margin of 0.61\% due to the introduced feature alignment using adversarial learning.
(2) As expected Joint Training outperforms both Off-the-Shelf and Finetuning baseline as it learns a shared representation from both dataset by transferring knowledge across them.

\begin{table}[t]
\caption{Experimental Results on \textbf{Single-label Object Recognition} task. 
The proposed approach, AKT outperforms all the unsupervised knowledge transfer methods.}
\label{table:task_1}
\begin{center}
\begin{tabular}{ l | c c }
\toprule[1.2pt]
\multicolumn{2}{c}{Target: CIFAR-10 and Source: CIFAR-100} \\
\toprule[1.2pt]
Methods & Target Accuracy (\%) \\ 
\bottomrule[1.2pt]
Scratch & 92.49 \\ 
\hline
Finetuning & 93.27 \\ 
Joint Training & 93.32 \\ 
\hline 
Pseudo Labels ~\cite{bachman2014learning}  & 92.85 \\
Random Network~\cite{pathak2016context} & 92.37 \\ 
Jigsaw~\cite{noroozi2016unsupervised} & 75.85 \\
Colorization~\cite{zhang2016colorful} & 92.57\\
Split-Brain~\cite{zhang2017split} & 92.60 \\
\hline
AKT (Ours: only $\mathbf{D}_I$) & 93.04 \\ 
AKT (Ours: with $\mathbf{D}_I$ and $\mathbf{D}_G)$ & \textbf{93.21} \\
\bottomrule[1.2pt]
\end{tabular}
\end{center}
\end{table}

\begin{table}[tb]
\caption{Comparison of our method with state-of-the-art self-supervised alternatives on \textbf{PASCAL-VOC Multi-label Object Classification} task. 
The reported results of all the self-supervised methods except the Rotation, Rotation Decoupling, and Pseudo Labels are from~\cite{noroozi2018boosting}}.
\begin{tabular}{ l | >{\centering\arraybackslash}p{3.2cm} }
\toprule[1.2pt]
\multicolumn{2}{c}{Target: PASCAL-VOC and Source: ImageNet} \\
\toprule[1.2pt]
Methods  & Target mAP (\%) \\
\bottomrule[1.2pt]
Scratch & 63.5 \\
\hline
Finetuning  & 87.0  \\
Joint Training & 86.7\\ 
\hline 
Pseudo Labels ~\cite{bachman2014learning}  & 63.2 \\
Random Network~\cite{pathak2016context} & 53.3  \\
Jigsaw~\cite{noroozi2016unsupervised} & 67.7 \\
Jigsaw++~\cite{noroozi2018boosting}& 69.9 \\
Colorization~\cite{zhang2016colorful} & 65.9 \\
Split-Brain~\cite{zhang2017split} & 67.1 \\
Rotation~\cite{gidaris2018unsupervised}& 73.0 \\
Rotation Decoupling~\cite{feng2019self}& 74.5 \\
\hline
AKT (Ours: only $\mathbf{D}_I$) & 76.9 \\
AKT (Ours: with $\mathbf{D}_I$ and $\mathbf{D}_G)$ & \textbf{77.4} \\
\bottomrule[1.2pt]
\end{tabular}
\label{table:task_5}
\end{table}

\subsubsection{Multi-label Object Recognition}
\quad We conduct this experiment on the more challenging multi-label PASCAL-VOC~\cite{everingham2010pascal} as the labeled target and ImageNet~\cite{deng2009imagenet} as the unlabeled source dataset. We follow the standard train/test split~\cite{everingham2010pascal} to perform our experiments. 
Table~\ref{table:task_5} shows the results. We have the following observations from Table~\ref{table:task_5}. (1) Our proposed method outperforms all other self-supervised methods including the Rotation Decoupling method in~\cite{feng2019self} by a significant margin (2.9\% improvement in mAP). 
Note that we train the source and target data jointly and since ImageNet is a large scale dataset of 1.1M images and PASCAL-VOC has only 4982 images, our method utilizes a small fraction of data from the ImageNet to achieve this state-of-the-art performance on the PASCAL-VOC dataset. (2) The performance gap between our method and supervised knowledge transfer methods begins to increase. This is expected as with a challenging multi-label object classification dataset, an unsupervised approach can not compete with a fully supervised knowledge transfer approach, especially, when the latter one is using true labels from a large scale source dataset like ImageNet. However, we would like to point out once more that, in practice, supervised knowledge transfer methods have serious issues with scalability as they require a tremendous amount of manual annotations. On the other hand, our approach can be trained on internet-scale datasets with no supervision. (3) The performance gap with Random Network baseline is much higher (53.3\% vs 77.4\%), justifying the relevance of our pseudo-label prediction using aligned features while transferring knowledge from a large unlabeled dataset. (4) Our proposed approach works the best while both instance-level and group-level discriminators are used for feature alignment (76.9\% vs 77.4\%). 

\begin{table}[t]
\begin{center}
\caption{Results on Font Character Recognition task.
Our approach outperforms all the self-supervised baselines and is very competitive against the supervised topline ($\sim$0.04\%). }
\label{table:task_2}
\
\begin{tabular}{ l | c c }
\toprule[1.2pt]
\multicolumn{2}{c}{Target: Char74K and Source: EMNIST} \\
\toprule[1.2pt]
Methods & Target Accuracy (\%) \\
\toprule[1.2pt]
Scratch & 8.55  \\
\hline
Finetuning & 19.85  \\
Joint Training & 18.60  \\
\hline 
Pseudo Labels ~\cite{bachman2014learning}  & 8.54\\
Random Network~\cite{pathak2016context} & 9.04  \\
Jigsaw~\cite{noroozi2016unsupervised} & 18.37 \\
\hline
AKT (Ours: only $\mathbf{D}_I$) & \textbf{19.81} \\
AKT (Ours: with $\mathbf{D}_I$ and $\mathbf{D}_G)$ & 19.79 \\
\bottomrule[1.2pt]
\end{tabular}
\end{center}
\end{table}

\hfill
\begin{table}[t]
\begin{center}
\caption{Experimental Results on \textbf{Handwritten Character Recognition}. Note that the performance of Random Network is very competitive for this task.}
\label{table:task_3}
\begin{tabular}{ l | c c }
\toprule[1.2pt]
\multicolumn{2}{c}{Target: EMNIST and Source: MNIST} \\
\toprule[1.2pt]
Methods & Target Accuracy (\%) \\ 
\bottomrule[1.2pt]
Scratch & 92.24 \\
\hline
Finetuning & 93.85 \\
Joint Training & 93.80 \\
\hline 
Pseudo Labels ~\cite{bachman2014learning}  & 89.16\\
Random Network~\cite{pathak2016context} & 92.35 \\
Jigsaw~\cite{noroozi2016unsupervised} & 50.90 \\
\hline
AKT (Ours: only $\mathbf{D}_I$) & 93.59 \\
AKT (Ours: with $\mathbf{D}_I$ and $\mathbf{D}_G)$ & \textbf{93.65}\\
\bottomrule[1.2pt]
\end{tabular}
\end{center}
\end{table}

\subsection{Character Recognition} 
The goal of this experiment is to compare our approach with other alternatives on recognizing both font and handwritten characters. 

\subsubsection{Font Character Recognition} 
\quad We use Char74K~\cite{de2009character}, a font character recognition dataset as the labeled target set and handwritten character split from EMNIST dataset \cite{cohen2017emnist} as the source dataset to perform this experiment. Char74K dataset consists of 74,000 images from 64 classes that includes English alphabets (a-z,A-Z) and numeric (0-9). 
EMNIST dataset consists of 145,600 images for 26 classes (a-z). 
For the Char74K dataset, we divide the data into 80/20 train/test split and use the standard train/test split for EMNIST. From Table~\ref{table:task_2}, we have the following observations. (1) Similar to the object recognition results, our proposed approach outperforms both Random Network and Pseudo Labels baseline by a significant margin (11\% improvement over Random Network). Among the alternatives, Jigsaw baseline is the most competitive. However, we still outperform the Jigsaw baseline by a margin of about 1.5\% in recognizing font characters. (2) Our approach is very competitive to the supervised knowledge transfer baselines (19.81\% vs 19.85\%) but significantly outperforms the training from scratch baseline by 11\%,
thanks to the effective knowledge transfer from unlabeled source data to the target dataset.

\subsubsection{Handwritten Character Recognition} 
~~We conduct this experiment using EMNIST as the labeled target dataset and MNIST~\cite{lecun1998mnist} as the unlabeled source dataset. While EMNIST contains handwritten characters, MNIST is a standard dataset for handwritten digits containing 80,000 images. 
We use the original train/test split from both datasets in our experiments. Table~\ref{table:task_3} shows that our proposed method achieves performance very close to the supervised methods which indicates that our method can achieve comparable performance without requiring a single label from the source dataset. Note that the performance of Jigsaw baseline is only 50.90\% which is much lower than that of training from scratch. We believe this is because the Jigsaw is optimized to work for $256 \times 256$ size images in training~\cite{noroozi2016unsupervised} and hence fails to learn efficient features with $28\times28$ small size images on this task.

\begin{table}[t]
\hfill
\begin{center}
\caption{Results on \textbf{Advertisement Dataset}. Our proposed approach performs the best among the self-supervised alternatives and is very competitive against supervised baselines.}

\label{table:task_4}
\begin{tabular}{ l | c }
\toprule[1.2pt]
\multicolumn{2}{c}{Target: Advertisement and Source: BAM} \\
\toprule[1.2pt]
Method & Target Accuracy (\%) \\ 
\toprule[1.2pt]
Scratch & 25.02  \\ 
\hline
Finetuning & 29.00  \\
Joint Training & 29.51  \\
\hline 
Pseudo Labels ~\cite{bachman2014learning} & 25.02 \\
Random Network~\cite{pathak2016context} & 25.51  \\
Jigsaw~\cite{noroozi2016unsupervised} & 28.38 \\
Colorization~\cite{zhang2016colorful} & 16.12 \\
Split-Brain~\cite{zhang2017split} & 27.71\\
\hline
AKT (Ours: only $\mathbf{D}_I$) & 28.70 \\
AKT (Ours: with $\mathbf{D}_I$ and $\mathbf{D}_G)$ & \textbf{28.87}\\
\bottomrule[1.2pt]
\end{tabular}
\end{center}
\end{table}

\subsection{Sentiment Recognition} 
Recently, sentiment analysis has garnered much interest in computer vision as there is more to images than their objective physical content: for example, advertisements are created to persuade a viewer to take a certain action. However, collecting manual labels for advertisement images are very difficult and costly compared to generic object labels. The goal of this experiment is to leverage unlabeled images for improving performance of a sentiment recognition classifier on advertisement images. We conduct this experiment using the Advertisement dataset~\cite{hussain2017automatic} as the labeled target and Behance-Artistic-Media (BAM)~\cite{wilber2017bam} dataset as the unlabeled source dataset. While the Advertisement dataset contains 64,382 images across 30 sentiment categories, the BAM dataset consists of 2.5 million images across 4 sentiment categories. 

Following are the observations from Table~\ref{table:task_4}: (1) Our method consistently outperforms the other baselines to achieve the top accuracy of 28.87\% on the Advertisement dataset. Jigsaw baseline is the most competitive with an accuracy of 28.38\%. We observe that Jigsaw baseline is able to transfer the knowledge well only if the pretext task is trained on a very large-scale dataset containing millions of images. (2) 
Since our algorithm processes same amount of the source and the target data, it uses much lower number of unlabeled samples from the source dataset. However, by utilizing some fraction of unlabeled samples, our approach is only about 0.13\% behind Finetuning baseline that uses all the unlabeled source samples along with their labels for training the source classifier.

\section{Ablation Analysis}
\begin{table*}[t]
    \caption{\textbf{Performance of proposed  method with different architecture on Object Recognition tasks.}}
    \begin{subtable}{.49\linewidth}
        \centering
       
        \begin{tabular}{ l | c c }
            \toprule[1.2pt]
            \multicolumn{2}{c}{Target: CIFAR-10 and Source: CIFAR-100} \\
            \toprule[1.2pt]
            Architecture & Target Accuracy (\%) \\
            \bottomrule[1.2pt]
            AlexNet~\cite{krizhevsky2012imagenet} &  89.04 \\
            VGG16~\cite{simonyan2014very} & 93.21\\
            ResNet50~\cite{he2016deep} &  94.65\\
            \bottomrule[1.2pt]
        \end{tabular}
         \caption{\textbf{Single-Label Object Recognition}}
         \label{tab:recog_s}
    \end{subtable}%
    \hspace{\fill}
    \begin{subtable}{.49\linewidth}
      \centering
      \
        \begin{tabular}{ l | c c }
            \toprule[1.2pt]
            \multicolumn{2}{c}{Target: PASCAL-VOC and Source: ImageNet} \\
            \toprule[1.2pt]
            Architecture & Target mAP (\%) \\
            \bottomrule[1.2pt]
            AlexNet~\cite{krizhevsky2012imagenet} &  77.4\\
            VGG16~\cite{simonyan2014very} & 78.2\\ 
            ResNet50~\cite{he2016deep} & 79.6\\
            \bottomrule[1.2pt]
        \end{tabular}
         \caption{\textbf{Multi-Label Object Recognition}}
         \label{tab:recog_m}
    \end{subtable} 
\end{table*}

\begin{figure*}[ht]
\label{fig:ablation_pseudo_labels}
\begin{subfigure}{0.47\textwidth}
\centering
    \includegraphics[width=\textwidth]{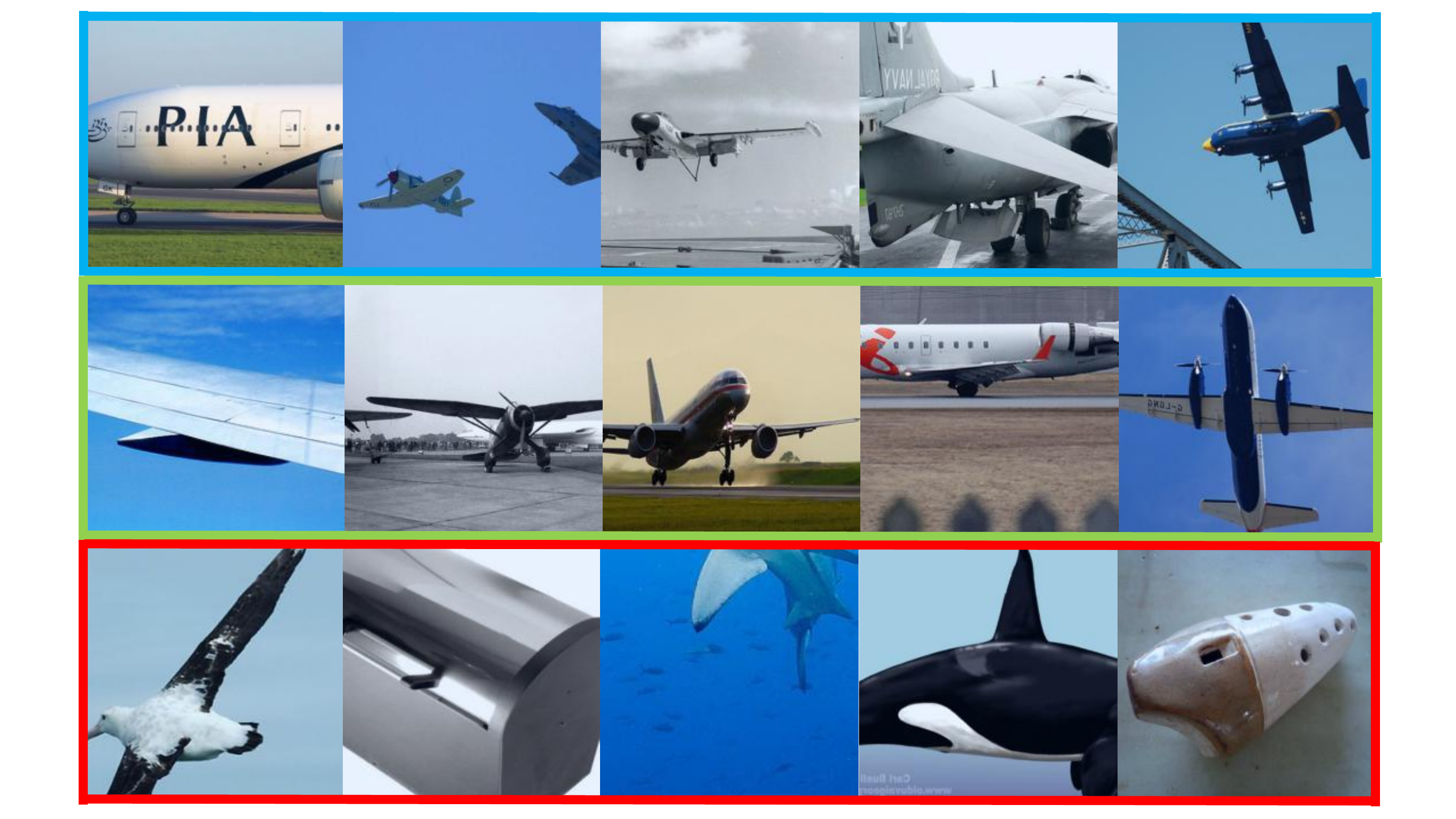}%
    \caption{Samples from class \textit{aeroplane} from PASCAL-VOC experiment.}
    \label{subfig:compare_a}
\end{subfigure}
\quad
\begin{subfigure}{0.47\textwidth}
\centering
    \includegraphics[width=\textwidth]{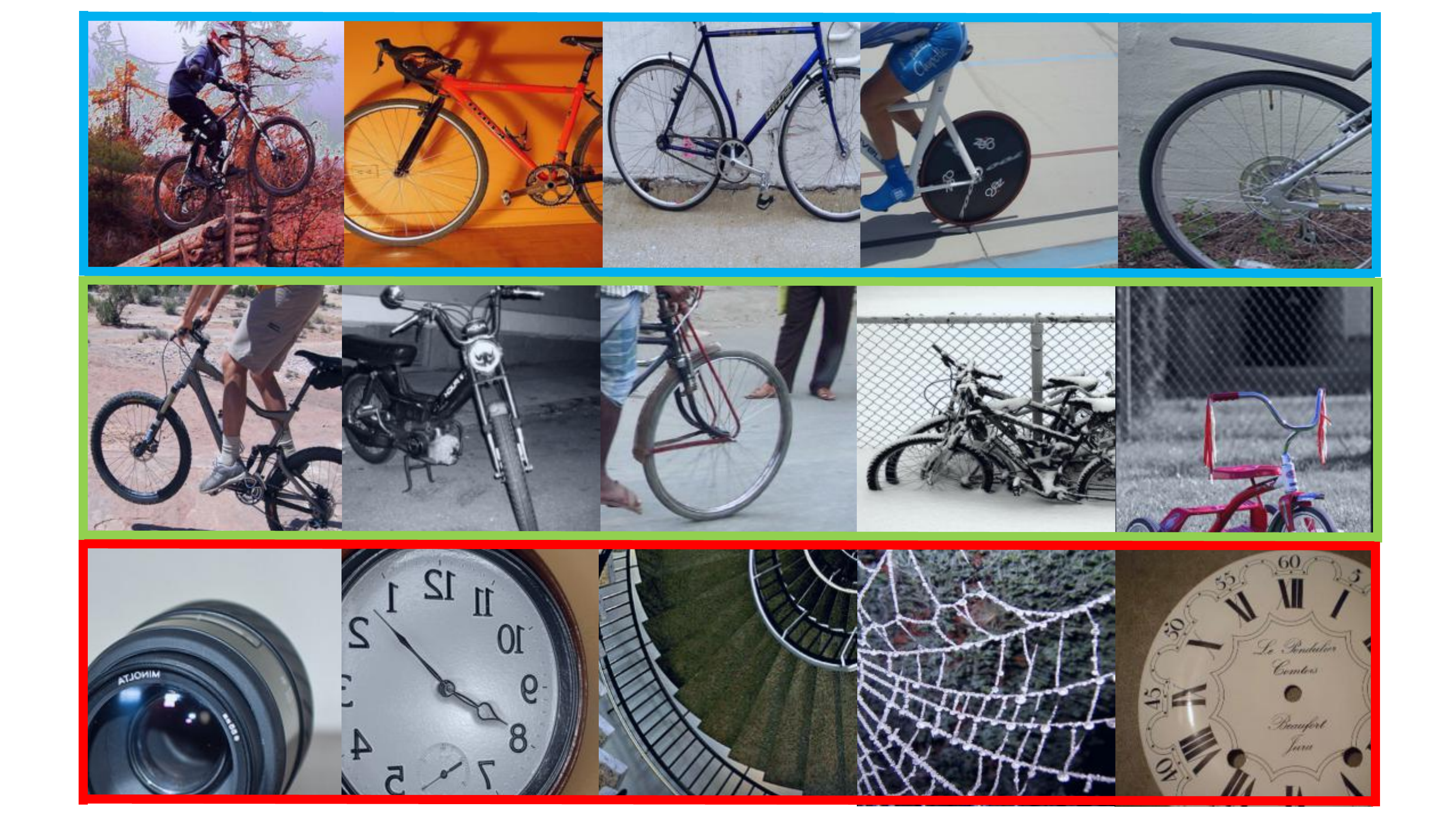}%
    \caption{Samples from class \textit{bike} from PASCAL-VOC experiment.}
    \label{subfig:compare_b}
\end{subfigure}
\caption{Predicted pseudo-labels on ImageNet data. Top-row (\textcolor{blue}{blue}) are the samples from the target PASCAL-VOC dataset. Middle-row (\textcolor{green}{green}) shows the samples from ImageNet dataset where pseudo labels correspond to correct labels. Bottom-row (\textcolor{red}{red}) shows the source samples where object in image does not match the correct label but has visual similarity to the target class. From the Middle-rows, we observe that our pseudo-label generator predicts the correct label of many unlabeled samples such as aeroplanes in (a) and bike in (b). Our approach classifies bird images (which resembles aeroplane) and fins of dolphins (which resembles wings of a aircraft) as aeroplanes as seen in bottom row of (a). Similarly, since target data in (b) have bike wheel as prominent feature, source samples with circular features are pseudo-labeled as bike. Best viewed in color.} 
\end{figure*}

We perform the following ablation experiments to better understand the contributions of various components of our proposed approach.

\subsection{Advantage of Feature Alignment} 
To better understand the contribution of feature alignment across target features from the classifier and source features from the pseudo-label generator, we perform an experiment where we jointly train the classifier with labeled target data and pseudo-labeled source data without any feature alignment. We observe that our approach without feature alignment produces inferior results, with target accuracy of 92.75\% on CIFAR-10/CIFAR-100 and 67.10\% on PASCAL-VOC/ImageNet experiments respectively. These performances are about 0.5\% and 10\% lower compared to our approach using feature alignment for CIFAR-10/CIFAR-100 and PASCAL-VOC/ImageNet experiments respectively. Thus, we conjecture that feature alignment is an essential step for predicting reliable pseudo-labels while transferring knowledge from the unlabeled data.

\subsection{Different Layer for Alignment} 
We examine the performance of our approach by using adversarial loss across fc6 instead of fc7 features along with conv5 features and found that feature alignment across fc6 produces inferior results (92.85\% vs 93.21\%) on CIFAR-10/CIFAR-100 experiment. This suggests that fc7 features along with conv5 features are better suited for transfer learning as shown in many prior works.

\subsection{Impact of Network Architecture} 
In the supervised setting, deeper architectures have shown  higher performance on many tasks. To study the impact of choice of architecture we perform experiments on object recognition tasks using VGG16 and ResNet50 architectures. As expected, we observe that a deeper architectures VGG16 and ResNet50 leads to significant improvement as compared to AlexNet.  From Table~\ref{tab:recog_s} we observe that on the CIFAR-10/CIFAR-100 task our proposed approach achieves 93.21\% with VGG16 and 94.65\% using ResNet50 as compared to 89.04\% on AlexNet. Similar performance trend is observed on multi-label recognition (Table~\ref{tab:recog_m}), where we achieve improvement of 0.8\% mAP and 2.2\% mAP using VGG16 and ResNet50, respectively.

\subsection{Reliability of Pseudo Labels}
From Figure~\ref{subfig:compare_a}-~\ref{subfig:compare_b}, we observe that the pseudo-label generator $\mathbf{G}$ is able to generate correct labels for many samples in the source dataset (middle row) and generates reasonable labels for visually similar classes (bottom row). We further compute reliability score by comparing the correct label and pseudo-labels of about 700 randomly sampled unlabeled images from ImageNet dataset and find that the reliability score is more than 85\% for both of the classes as shown in Table~\ref{tab:reliable}.

\subsection{MSE Loss vs Adversarial Loss}
We investigate the importance of adversarial loss by comparing with $L_{2}$ loss for aligning the features on the CIFAR-10/CIFAR-100 task, and found that the later produces inferior results with a target accuracy of 92.66\% compared to 93.21\% by the adversarial loss.

\subsection{Out-of-Distribution Source Data}
Following~\cite{guo2019new}, we perform an experiment using mini-ImageNet as the unlabeled source dataset and EuroSAT containing remote sensing images with no perspective distortion, as the labeled target dataset. Our proposed approach outperforms the Pseudo-Labels baseline by a margin of 0.72\% (95.22\% vs 94.5\%) and is competitive with the supervised Fine-tuning baseline (95.22\% vs 96.19\%) using AlexNet as the network architecture. 

\begin{table}[ht]
    \centering
    \caption{Top-3 pseudo label predictions on ImageNet classes.} 
    \renewcommand{\arraystretch}{1.2}
    \begin{tabular}{l|l|c}
        \toprule[1.2pt]
         ImageNet Class & Top-3 Pseudo Label & Score\\
         \toprule[1.2pt]
         Fig.~\ref{subfig:compare_a}. warplane & \textbf{aeroplane}, bird, car & 86.67\% \\
         Fig.~\ref{subfig:compare_b}. bike & \textbf{bicycle}, motorbike, person & 88.46\% \\
         \bottomrule[1.2pt]
    \end{tabular}
    \label{tab:reliable}
\end{table}

\section{Conclusion}

We present an Adversarial Knowledge Transfer (\textbf{AKT}) approach for transferring knowledge from unlabeled data to the labeled data without requiring the unlabeled data to be from the same label space or data distribution as of the labeled data. The proposed adversarial learning jointly trains the classifier using both labeled target data and source data whose labels are predicted using a pseudo-label generator by aligning the feature space of unlabeled data with the labeled target data. Experiments show that our approach not only outperforms the unsupervised knowledge transfer alternatives but is also very competitive while comparing against supervised knowledge transfer methods.

\begin{acks}
This work is partially supported by NSF grant 1724341, ONR grant N00014-18-1-2252 and gifts from Adobe. We thank Abhishek Aich for his assistance and feedback on the paper.
\end{acks}

\bibliographystyle{ACM-Reference-Format}
\bibliography{sample-base}

\appendix


\end{document}